\documentclass[letterpaper, 10 pt, conference]{ieeeconf}  

\IEEEoverridecommandlockouts                              

\usepackage{graphics} 
\usepackage{epsfig} 
\usepackage{mathptmx} 
\usepackage{times} 
\usepackage{amsmath} 
\usepackage{multirow}
\usepackage{amssymb}  
\usepackage{color}
\usepackage{gensymb}
\usepackage{siunitx}
\usepackage{lipsum}
\usepackage{mathtools, nccmath}
\usepackage{cuted}
\usepackage{hhline}
\usepackage{makecell}
\usepackage{caption}
\usepackage{cite}
\usepackage[subtle,bibbreaks=normal,paragraphs=tight,floats=normal,mathspacing=normal,wordspacing=normal,tracking=normal]{savetrees}

\usepackage{hyperref}
\hypersetup{
    colorlinks=true,
    linkcolor=blue,
    filecolor=magenta,      
    urlcolor=blue,
}

\newcommand{\myparagraph}[1]{\vspace{0.05in}\noindent\textbf{#1}}

\DeclareMathOperator*{\argmin}{arg\,min}

\title{\bf
Simultaneous Tactile Estimation and Control of Extrinsic Contact
\vspace{-1ex}}
\author{
  \authorblockN{Sangwoon Kim$^{1}$, 
Devesh K. Jha$^{2}$, Diego Romeres$^{2}$, Parag Patre$^{3}$ and Alberto Rodriguez$^{1}$} 
  \authorblockA{
     $^{1}$MIT, $^{2}$Mitsubishi Electric Research Laboratories, $^{3}$Magna International Inc.\\
     {\tt\small <sangwoon,albertor>@mit.edu, <jha,romeres>@merl.com, parag.patre@magna.com}} \
\vspace{-5ex}
}

\begin{document}

\maketitle

\thispagestyle{empty}
\pagestyle{empty}


\begin{abstract}

We propose a method that simultaneously estimates and controls extrinsic contact with tactile feedback. The method enables challenging manipulation tasks that require controlling light forces and accurate motions in contact, such as balancing an unknown object on a thin rod standing upright. A factor graph-based framework fuses a sequence of tactile and kinematic measurements to estimate and control the interaction between gripper-object-environment, including the location and wrench at the extrinsic contact between the grasped object and the environment and the grasp wrench transferred from the gripper to the object. The same framework simultaneously plans the gripper motions that make it possible to estimate the state while satisfying regularizing control objectives to prevent slip, such as minimizing the grasp wrench and minimizing frictional force at the extrinsic contact. We show results with sub-millimeter contact localization error and good slip prevention even on slippery environments, for multiple contact formations (point, line, patch contact) and transitions between them. See supplementary video and results at \href{https://sites.google.com/view/sim-tact}{https://sites.google.com/view/sim-tact}.

\end{abstract}

\section{Introduction}\label{sec:introduction}

Tactile feedback is specially useful in manipulation tasks where visual information is limited, or tasks that involve contact with an unknown or uncertain object and environment~\cite{Rodriguez}. We are specially interested in tasks where precise contact regulation is important, but at the same time difficult to observe. 

\begin{figure}[t]
	\centering
	\includegraphics[width=\linewidth]{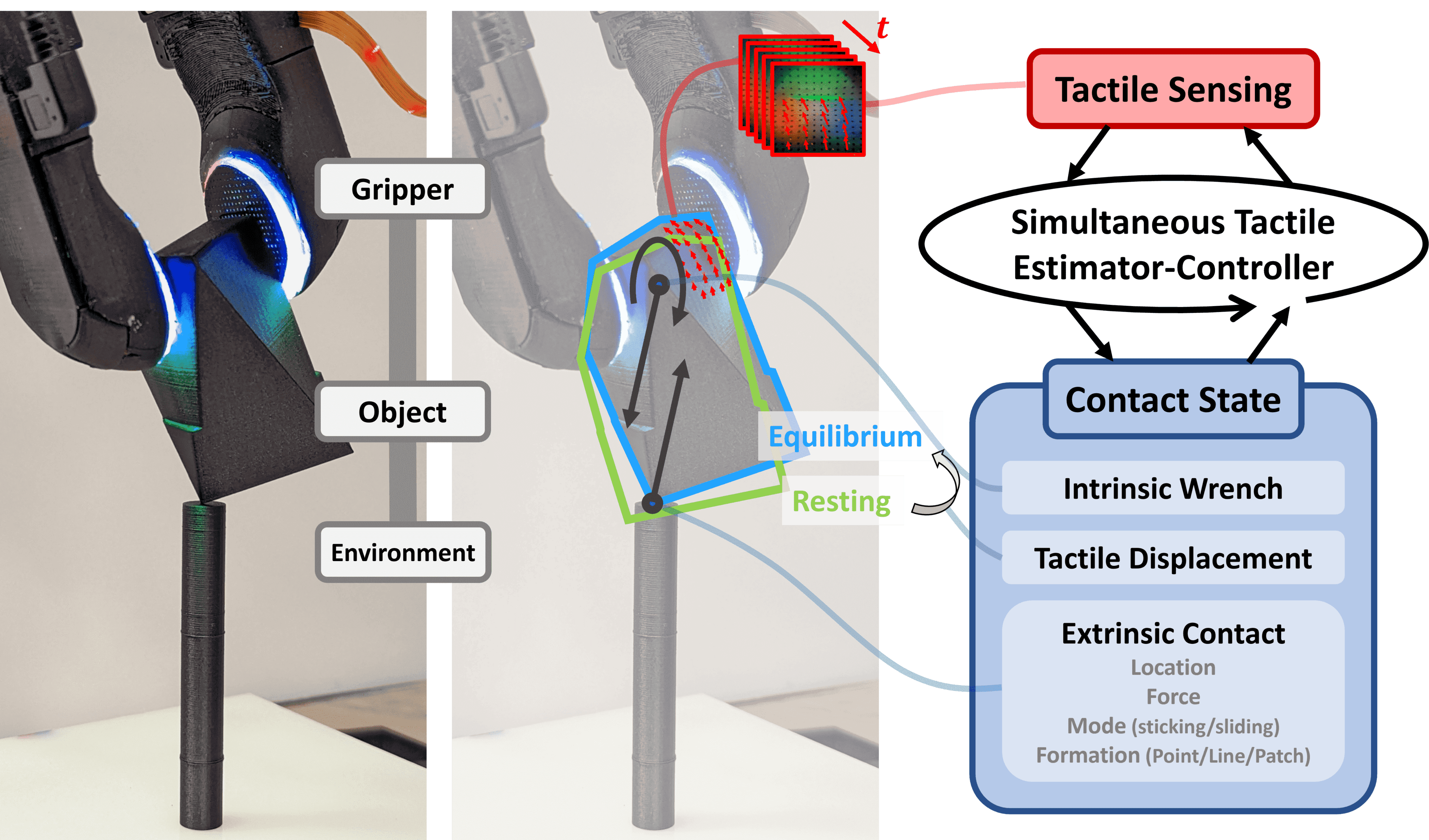}
	\caption{Simultaneous tactile estimator-controller used to stably place an unknown object on an unsupported thin rod standing upright.}
	\label{fig:overview}
	\vspace{-6mm}
\end{figure}

Figure~\ref{fig:overview} shows an illustrative--maybe extreme--example. A gripper is holding an unknown object and is attempting to balance a corner of that object on a free-standing rod. Forces that are transmitted directly vertically at the contact between the object and the rod (\textit{extrinsic contact}) will excite the internal forces of the grasp (\textit{intrinsic contact}) and generate informative signals for estimating the location of that contact. But forces in any other direction are to be avoided, they will make the rod pivot and quickly tumble. When the rod is thin, this is a very difficult--close to impossible--task for a person to control, even with direct line of sight, requiring both compliance and kinematic precision.




In this work we show that, by integrating tactile and kinematic measurements, and with a simultaneous estimation and control framework, it is possible for a robot to do it blind, with arbitrary objects, and with poor prior information of where the external contact is located. Realizing this simultaneous estimation-control behavior with tactile feedback involves the following challenges:
\begin{itemize}
    \item A single snapshot from tactile sensors does not fully describe the contact configuration or the kinematic state of the system~\cite{Newman}. We need to fuse measurements over time along informative motions to estimate it with some certainty.
    \item The extrinsic contact between the grasped object and the environment is un-sensed, and to reason about it we need to infer through the chain of contacts that connects the end-effector to the object and to the environment. The tactile sensors we use allow to directly observe the compliance between the object and the end-effector, and tracking that compliance makes it possible to reason about the behavior of the extrinsic contact (as in \cite{Ma2021}). 
\end{itemize}

This paper proposes a framework that uses a factor graph-based simultaneous tactile estimator-controller framework to: 1) \textbf{estimate} the contact state: object's relative displacement, extrinsic contact location and formation, and the wrench exerted at the intrinsic and extrinsic contacts; and  simultaneously, 2) \textbf{control} the contact state to the desired configuration. We especially focus on localizing the contact with active motion while maintaining sticking contact between object and environment in multiple contact formations, importantly, doing this while minimizing the forces exerted on the environment.

\section{Related Work}\label{sec:related_work}

\begin{figure*}[t]
	\centering
	\includegraphics[width=0.95\linewidth]{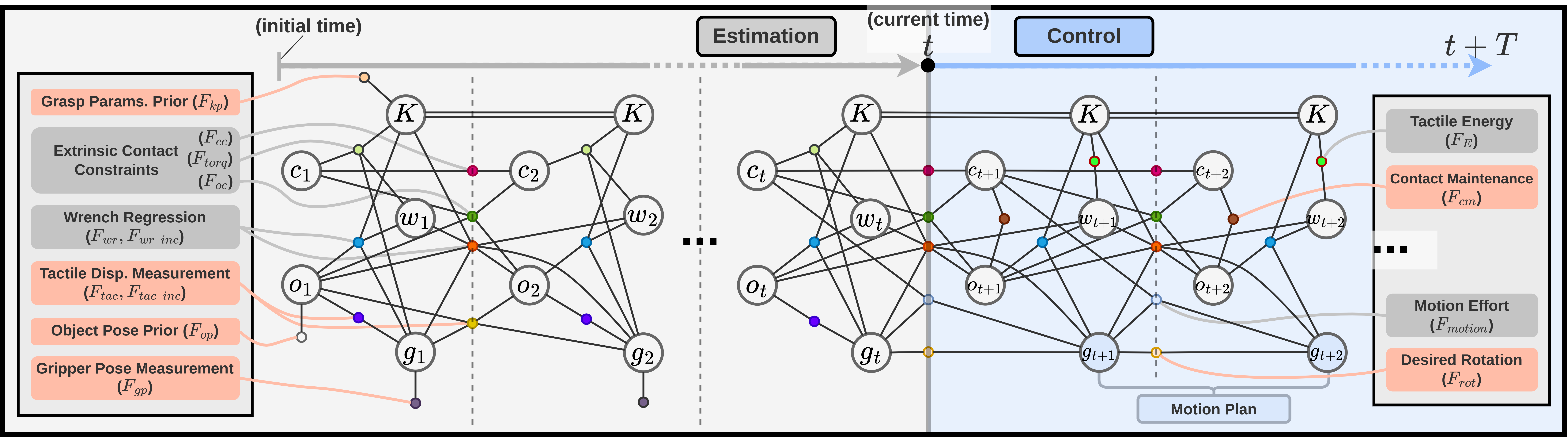}
	\vspace{-1.5mm}
	\caption{Factor graph architecture for the simultaneous tactile estimator-controller}
	\label{fig:graph}
	\vspace{-6.5mm}
\end{figure*}

\subsection{Contact State Estimation and Control}

There has been prior work on contact state estimation and control using tactile sensors \cite{Yuan, Donlon, Taylor, wang2022tacto, Lambeta}. One line of the work is about avoiding slipping and maintaining sticking contact at the intrinsic contact between the sensor finger and the grasped object \cite{veiga2015stabilizing, li2014learning, Dong2019ICRA}. While maintaining the sticking contact, other researchers use tactile sensing to control the pose of the manipulated objects when executing manipulation primitives \cite{Hogan, tian2019manipulation}. There also has been research that studies sliding dynamics at the gripper finger \cite{She}. Lastly, researchers have used tactile sensing for planning and decision-making \cite{kolamuri2021improving, Dong2021}. However, most of these methods focus only on the intrinsic contact state and do not reason carefully about the extrinsic contact state.

Doshi et al. \cite{doshi2022manipulation} developed a contact configuration estimator and controller to manipulate unknown objects. Their task and approach are similar to ours in that they do both estimation and control. They used force-torque sensing instead of tactile sensing, which eased a challenge mentioned in Sec.\ref{sec:introduction}. Kim and Rodriguez \cite{Kim2021} used tactile sensors to localize the extrinsic contact and solved the peg-in-hole insertion using the estimated contact location. They used a proportional controller to maintain the constant intrinsic finger wrench in order to avoid slippage at extrinsic contact. While both above methods estimate and control the extrinsic contact, they do them separately in independent architectures rather than simultaneously in an integrated architecture.

\subsection{Factor Graph for Estimation and Control}

A factor graph is a bipartite graphical model composed of variables and factors where each factor represents a function on a subset of the variables. One application recently gaining attention is using factor graphs for planning and control \cite{Dellaert2021}. For example, one can construct a linear quadratic regulator (LQR) by representing the state and control input as variables and system dynamics as factors \cite{Yang2021}. Dong et al. \cite{Dong2016} formulate the Gaussian Process trajectory prior as a chain of variables and factors to do motion planning that results in a smooth trajectory. The same researchers demonstrate simultaneous motion planning (future) and trajectory estimation (past) in the same factor graph \cite{Mukadam2019}. This is especially relevant to our work, where our framework also tries to control (future) and estimate (past) the contact state simultaneously.

\section{Problem Formulation}

The problem we solve is to estimate and control the contact state in multiple contact formations and to detect the transition between the different contact formations. The conditions of the problem are:

\begin{itemize}
    \item The object is unknown and rigid. The bottom of the object has appropriate shape to make it possible to be placed stably (e.g., corners, edges, flat bottom). 
    \item The environment is flat at the contact with known normal direction.
    \item At the first contact, the object and the environment meet at a point contact.
    \item We do not make use of visual feedback.
\end{itemize}

\noindent Specifically, we would like to localize and control the contact between object and environment with minimal slip by minimizing the tangential force at extrinsic contact. To do so, we estimate and control the contact state shown in Fig.\ref{fig:overview}:

\begin{itemize}
    \item gripper-object relative displacement from resting pose to equilibrium pose (tactile displacement)
    \item location and formation of extrinsic contact
    \item intrinsic wrench exerted from the gripper to object, and a resultant extrinsic contact force
\end{itemize}

\noindent The state is composed by five quantities:

\begin{itemize}
    \item $\mathbf{g} \in SE(3)$ - gripper pose
    \item $\mathbf{o} = \{{}^{r}o, {}^{eq}o\} \in SE(3)$ - object pose
    \item $\mathbf{w} = \{M_x,M_y,M_z,F_x,F_y,F_z\} \in \mathbb{R}^6$ - intrinsic wrench
    \item $\mathbf{c} \in SE(3)$ - contact pose
    \item $\mathbf{K} = \{\kappa_x, \kappa_y, \kappa_z, k_x, k_y, k_z, \eta_x, \eta_y, \eta_z\} \in \mathbb{R}^9$ - grasp parameters
\end{itemize}

\noindent ${}^{r}o$ and ${}^{eq}o$ are the resting and equilibrium object poses. $\mathbf{w}$ is the wrench the gripper exerts on the object. $\kappa$ and $k$ are the rotational and translational stiffness of the grasp due to the elasticity of the gripper finger. $\eta$ is the offset of the center of compliance from the gripper center point. The center of compliance is the point where there is no translational displacement when pure torque is exerted on the object, and also it is where we define the intrinsic wrench ($w$). $\kappa$, $k$, and $\eta$ are combined as $K$, which we call ``grasp parameters". The grasp parameters vary depending on the local geometry of the object and the grasping force. 

For a point contact, we only need translational components of the contact pose, which makes the rotation of $\mathbf{c}$ redundant. To handle this redundancy, we fix the rotation to be aligned with the environment frame. Similarly, we only need one rotational dimension for line contact since we already have that the contact line is on the environment surface. Therefore, we constrain the z-axis of $\mathbf{c}$ to be perpendicular to the environment surface, while the x-axis represents the contact line.

\section{Simultaneous Tactile Estimator-Controller}

We use a factor graph to simultaneously estimate and control the contact state with tactile feedback, shown in Fig.\ref{fig:graph}. Each circle represents a variable, and each dot represents a factor. Factors with the red labels take input as measurements, prior, or commands.
The left part of the architecture, from timestep 1 to $t$ (past), is responsible for aggregating measurements and estimating the past/current contact states. The part from timestep $t+1$ to $t+T$ (future) is accountable for turning the desired rotation of the object into planned gripper motion (blue circles) and predict the future contact states. $T$ is the length of the control horizon. Both parts can be solved simultaneously by minimizing the total factor costs, which breaks down to a nonlinear least-squares problem:

\vspace{-2mm}\begin{align}
    \hat{x} = \argmin_{x} \sum_f ||F_f(x_{1:t}, x_{t+1:t+T}; z_{1:t},  z_{t+1:t+T})||_{\sum_f}^2
    \label{eq:squares}
    %
\end{align}\vspace{-5mm}

\noindent where $x_{1:t}$ and $x_{t+1:t+T}$ are past and future states, $z_{1:t}$ is input from measurements and priors, and $z_{t+1:t+T}$ is input from commands. $F_f$ represents factors in Fig.\ref{fig:graph}. $\sum_f$ is the covariance of the factors' noise model.

When the new measurement arrives, the factors at timestep $t+1$ are modified, and the new measurement is added, so the border between the estimation and control part is pushed one step forward. Also, one control step is added at $t+T+1$, so the length of the control horizon remains constant at $T$. Then, we solve this modified least-squares problem to update the solution. We use an incremental solver \cite{Kaess2012}, which allows fast computation. After each update, the gripper motion plan is sent to the robot, and the robot follows that trajectory until the next update. The update is asynchronous from the robot control; therefore, the robot does not have to wait until the next update unless it reaches the end of the previous motion plan trajectory.

\begin{figure*}[t]
	\centering
	\includegraphics[width=0.8\linewidth]{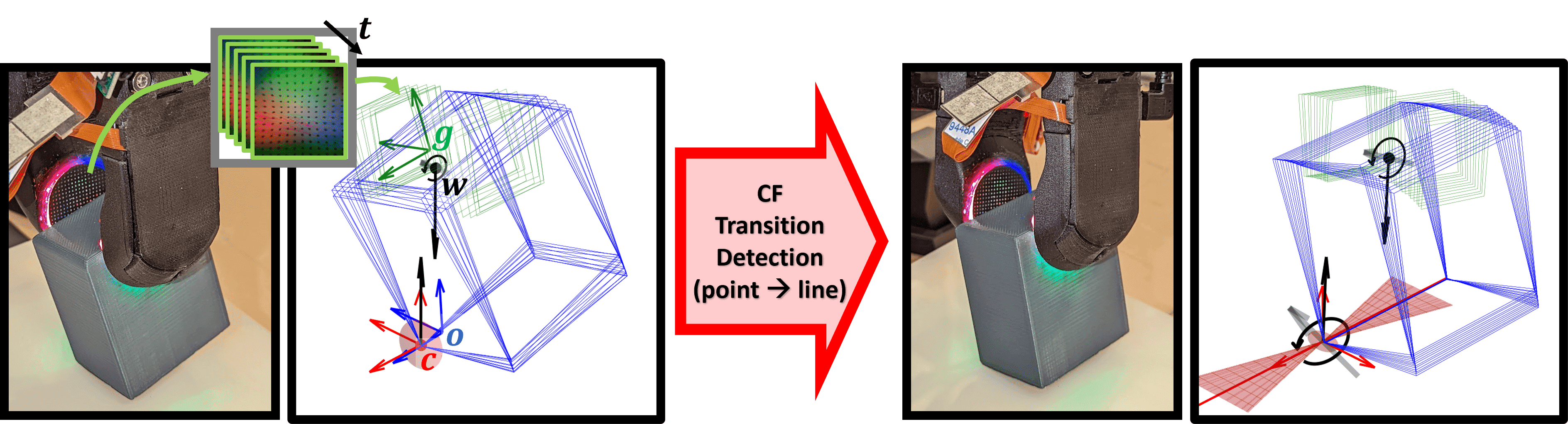}
	\vspace{-1mm}
	\caption{An object making point (left) and line (right) contact with the horizontal environment}
	\label{fig:viz}
	\vspace{-6mm}
\end{figure*}

\begin{figure}[t]
	\centering
	\vspace{-1mm}
	\includegraphics[width=\linewidth]{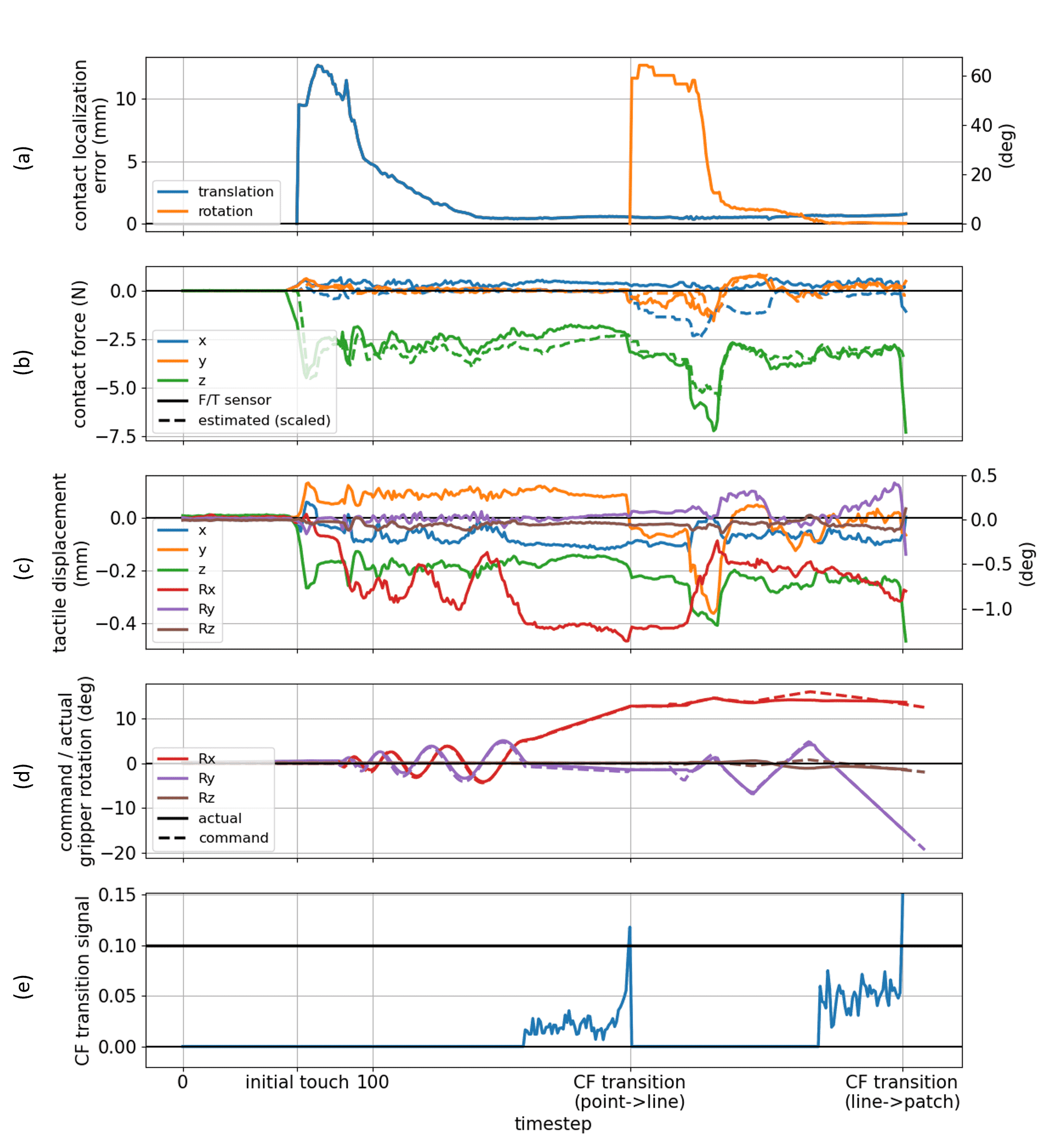}
	\vspace{-6mm}
	\caption{Time series plots for the case in Fig.\ref{fig:viz}}
	\label{fig:plots}
	\vspace{-7mm}
\end{figure}

\subsection{Contact State Estimation}

We describe the terms in Eq.\ref{eq:squares} that are responsible for contact state estimation ($\hat{x}_{1:t}$). These terms take inputs from priors and measurements and impose costs for constraints.

\myparagraph{Priors}: Unary factors impose priors on the variables.

\vspace{-2mm}\begin{align}
    F_{kp}(K;K^*) &:= K - K^*\\
    F_{op}(o_1;o^*) &:= o^{*-1} ({}^{r}o_1)
\end{align}\vspace{-6mm}

\noindent$K^*$ and $o^*$ are the priors for grasp parameters and initial object pose. We obtain grasp parameters prior by fitting it to one randomly selected grasp on the rectangular test object. Then, the same unmodified prior is used for other grasps/objects. A weak initial object pose prior is selected to have a bottom surface parallel to the environment because we do not know object shape and orientation.

\myparagraph{Gripper Pose Measurement}: The gripper pose measurement from forward kinematics ($g_i^*$) is imposed as a unary factor:

\vspace{-1.5mm}\begin{align}
    F_{gp}(g_i;g_i^*) &:= g_i^{*-1} g_i
\end{align}\vspace{-5.5mm}

\myparagraph{Tactile Displacement Measurement}: To measure the displacement of the object from resting to equilibrium pose under contact, we use GelSlim 3.0, a vision-based tactile sensor that observes the deformation of the sensor finger due to contact as a high-resolution tactile image. We feed the tactile image to the tactile module adopted from \cite{Kim2021} to get the object displacement ($\delta_i$). The tactile module is a convolutional neural network that takes the tactile image as input and outputs the tactile displacement. We use two types of factors to impose the measurement: one for total displacement ($F_{tac}$) and the other for incremental displacement ($F_{tac\_inc}$):

\vspace{-2mm}\begin{align}
    &F_{tac}(g_i,o_i;\delta_i) = \delta_i^{-1} (({}^{r} o_i^{-1} g_i)^{-1} ({}^{eq} o_i^{-1} g_i))\\
    &F_{tac\_inc}(g_{i-1},o_{i-1},g_i,o_i;\delta_{i-1},\delta_i) \nonumber\\
    &\hspace{1cm}= (\delta_{i-1}^{-1} \delta_i)^{-1} (({}^{eq} o_{i-1}^{-1} g_{i-1})^{-1} ({}^{eq} o_i^{-1} g_i))
\end{align}\vspace{-6mm}

\myparagraph{Extrinsic Contact Geometric Constraints}:

\vspace{-2mm}\begin{align}
    F_{oc}(o_{i-1},c_{i-1}o_i,c_i) &= ({}^{eq}o_{i-1}^{-1}c_{i-1})^{-1}({}^{eq}o_i^{-1}c_i)\\
    F_{cc}(c_{i-1},c_i) &= c_{i-1}^{-1} c_i
\end{align}\vspace{-6mm}

\noindent \textbf{$\mathbf{F_{oc}}$}: Unless there is a transition in the contact formation, the location of contact in the object frame should remain constant. We impose it as a strong factor between adjacent timesteps.
\noindent \textbf{$\mathbf{F_{cc}}$}: Since we assume a flat environment, the location of contact on the environment should not change in the direction perpendicular to the environment surface. Also, the change in the tangential direction should be small if the control objective of minimizing slip is met properly. We impose this by formulating a binary factor and setting a strong cost in the perpendicular direction and less strong in the tangential direction.

\myparagraph{Extrinsic Contact Torque Constraints}: During point contact, there should not be torque exerted about the contact, and during line contact, there should not be torque component in the direction parallel to the contact line.

\vspace{-2mm}\begin{align}
    &\quad {}^{p}F_{torq}(g_i,w_i,c_i,K) = \Vec{M} - \Vec{r}(g_i,c_i,\eta) \times \Vec{F} \nonumber\\
    &\qquad = \Vec{M} - ((g_i^{-1}c_i)_{trn}-\eta) \times \Vec{F} \quad \text{(point contact)} \nonumber\\
    &\quad {}^{l}F_{torq}(g_i,w_i,c_i,K) = {}^{p}F_{torq} \cdot \Vec{a}_x(g_i,c_i)  \quad \text{(line contact)}
    \label{eq:torq}
\end{align}\vspace{-6mm}

\noindent where $(g_i^{-1}c_i)_{trn}$ is the translational location of the contact with respect to the gripper, and $\Vec{a}_x(g_i,c_i)$ is the unit vector parallel to the estimated contact line.

\myparagraph{Wrench Regression Constraints}: To estimate the intrinsic wrench from the measurements, we approximate the grasp as decoupled linear springs in each rotational and translational direction, with an additional nonlinear term $\Delta$:

\vspace{-2mm}\begin{align}
    w_i = [\ \Vec{M}, \ \Vec{F} \ ] = [\ \mathbf{\kappa}, \ \mathbf{k} \ ] \odot({}^{r} o_i^{-1} g_i)^{-1} ({}^{eq} o_i^{-1} g_i) + \Delta
    \label{eq:wrench}
\end{align}\vspace{-6mm}

\noindent where $\odot$ is element-wise multiplication. $({}^{r} o_i^{-1} g_i)^{-1} ({}^{eq} o_i^{-1} g_i)$ is the tactile displacement in a canonical coordinate $[R_x,R_y,R_z,x,y,z]$, which we dropped the coordinate notation for simplicity. We regress the wrench close to the linear relation by imposing two types of factors:

\vspace{-2mm}\begin{align}
    &F_{wr}(g_i,o_i,w_i,K) = w_i - [\mathbf{\kappa}, \mathbf{k}] \odot({}^{r} o_i^{-1} g_i)^{-1} ({}^{eq} o_i^{-1} g_i) \\
    &F_{wr\_inc}(g_{i-1},o_{i-1},w_{i-1},g_i,o_i,w_i,K) \nonumber\\
    &\ = (w_i - w_{i-1}) - [\ \mathbf{\kappa}, \ \mathbf{k} \ ] \odot ({}^{eq} o_{i-1}^{-1} g_{i-1})^{-1} ({}^{eq} o_i^{-1} g_i)
\end{align}\vspace{-6mm}

\noindent $F_{wr}$ penalize the additional nonlinear term. $F_{wr\_inc}$ tries to regress the incremental change in the intrinsic wrench to be parallel to the linear relation.

\subsection{Contact State Control}

We describe the terms in Eq.\ref{eq:squares} that are responsible for contact state control and prediction ($\hat{x}_{t+1:t+T}$). In a nutshell, the system achieves the behavior of pivoting about an unknown contact point or line as the combination of these objectives: 1) A desired rotation of the object; 2) Minimizing motion effort and tactile deformation; 3) Maintaining a minimum contact with the environment.

\myparagraph{Desired Rotation}: While the factor graph computes the fine motion, we still need a input command in which direction we want to rotate (tilt) the object:

\vspace{-2mm}\begin{align}
    F_{rot}(g_{i-1},g_i;R_i) = R_i^{-1} (g_{i-1}^{-1} g_i)_{rot}
\end{align}\vspace{-6mm}

\noindent where $R_i$ is the desired rotation. This factor is tuned to be weaker than other control objectives, so the actual rotation can deviate from the desired rotation if this factor conflicts with other control objective factors (e.g., \textbf{tactile energy}). For example, in Fig.\ref{fig:plots}d, the actual rotation (solid line) deviates from the desired rotation (dashed line).

\myparagraph{Motion Effort}: We impose a cost on the local motion at the estimated contact point:

\vspace{-2mm}\begin{align}
    F_{motion}(g_{i-1},g_i,c_{i-1}) = c_{i-1}^{-1} g_i g_{i-1}^{-1} c_{i-1}
\end{align}\vspace{-6mm}

\myparagraph{Tactile Energy}: As we regress the intrinsic wrench to the decoupled linear relation, we approximate the elastic potential energy on the sensor finger as the quadratic sum of wrench components and impose it as a factor:

\vspace{-2mm}\begin{gather}
    E = \frac{1}{2} \{\frac{M_x^2}{\kappa_x}+\frac{M_y^2}{\kappa_y}+\frac{M_z^2}{\kappa_x}+\frac{F_x^2}{k_x}+\frac{F_y^2}{k_y}+\frac{F_z^2}{k_z}\}\\
    \Rightarrow F_{E}(w_i,K) = w_i \oslash [ \ \sqrt{\mathbf{\kappa}}, \ \sqrt{\mathbf{k}} \ ]
\end{gather}\vspace{-6mm}

\noindent where $\oslash$ is element-wise division.

This factor plays two important roles. First, it enables the prediction of future tactile displacement. This is because, given the extrinsic contact as a constraint, the intrinsic contact will be at a stable state where it minimizes the potential energy. Also, this factor helps to achieve control objectives while having minimal deformation on the sensor finger.

\myparagraph{Contact Maintenance}: None of the above control objective factors force the robot to push the object against the environment. In fact, minimizing \textbf{tactile energy} encourages moving away from the environment. Therefore, we impose a factor that encourages the robot to push the object to the environment by setting the target contact an offset distance inside the environment ($\epsilon_i$). We use a hinge function that penalizes when the estimated offset distance is less than the desired distance:

\vspace{-2mm}\begin{align}
    &F_{cm}(o_i,c_i;\epsilon_i) = \max(0, \zeta_i(o_i,c_i) - \epsilon_i ) \nonumber\\
    &\ = \max(0, -(({}^{r}o_i^{-1} c)^{-1}({}^{eq}o_i^{-1} c))_{trn,z} - \epsilon_i)
\end{align}\vspace{-6mm}

\noindent where $\zeta_i$ is the estimated offset distance.

When this factor is combined with the \textbf{tactile energy} factor, it enables minimizing the tangential force at the extrinsic contact. Imagine the contact state with no tangential force at the extrinsic contact. If we move the gripper parallel to the environment surface by applying tangential force in any direction, it will do positive work, which adds to the tactile energy while maintaining the constant offset distance. In other words, if the offset distance is fixed, minimizing the tactile energy is equivalent to minimizing the tangential force.

\subsection{Detection of Contact Formation Transition}

When the contact formation transitions from one to another, the \textbf{extrinsic contact torque constraints} are violated. Therefore, the transition can be detected by simply measuring an increase in the residual in Eq.\ref{eq:torq}. Fig.\ref{fig:plots}e shows an example of detecting transitions from point-to-line and line-to-patch, where the horizontal line at 0.1 is the threshold for the detection. After the transition, we can continue to use the estimator-controller by modifying the noise model of the factors accordingly.

\section{Experiments and Results}

\begin{figure}[t]
	\centering
	\vspace{1mm}
	\includegraphics[width=0.7\linewidth]{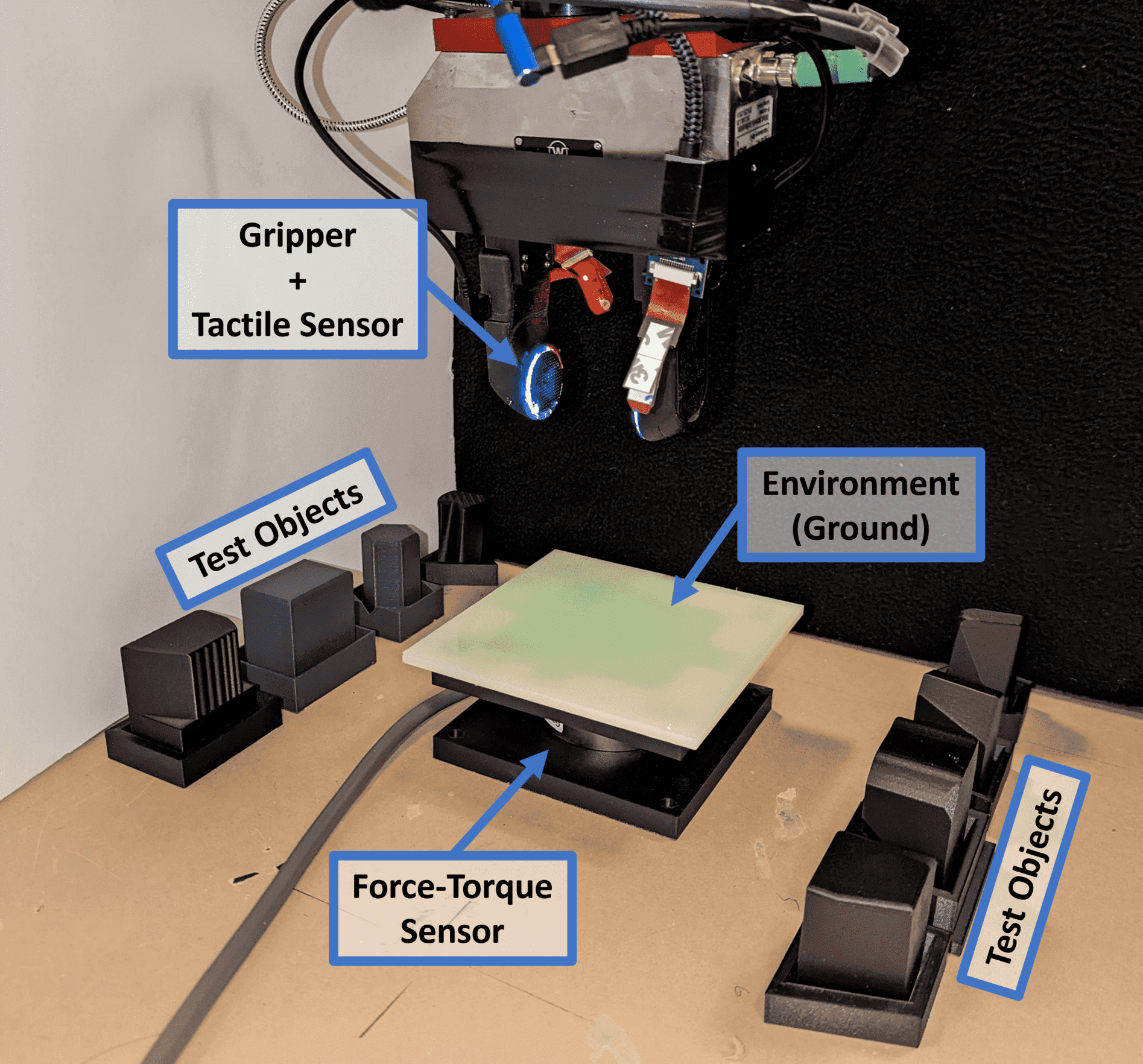}
	\vspace{-2mm}
	\caption{Experimental setup}
	\label{fig:setup}
	\vspace{-2mm}
\end{figure}

\setlength{\tabcolsep}{3pt} 
\renewcommand{\arraystretch}{1.5}
\begin{table*}[t]
\caption{Evaluation of Simultaneous Tactile Estimator-Controller on Point Contact Formation}\label{table:point}
\vspace{-2mm}
\begin{tabular}{|c||cccccccccc||cccccc}
\hline
\multirow{4}{*}{} & \multicolumn{10}{c||}{Higher Friction (Paper)}                                                                                                                                                                                                                                                                                                                                                                                                                                                                                                                                                                                                                                                                                                                                                                                    & \multicolumn{6}{c|}{Lower Friction (Acrylic)}                                                                                                                                                                                                                                                                                                                                                                                                                                                                                 \\ \cline{2-17} 
                  & \multicolumn{5}{c|}{Rectangle}                                                                                                                                                                                                                                                                                                                                                                                            & \multicolumn{5}{c||}{Hexagon}                                                                                                                                                                                                                                                                                                                                                                         & \multicolumn{3}{c|}{Rectangle}                                                                                                                                                                                                                                & \multicolumn{3}{c|}{Hexagon}                                                                                                                                                                                                                                  \\ \cline{2-17} 
                  & \multicolumn{1}{c|}{\multirow{2}{*}{\begin{tabular}[c]{@{}c@{}}loc.\vspace{-4pt}\\ error\vspace{-4pt}\\ (mm)\end{tabular}}} & \multicolumn{2}{c|}{\begin{tabular}[c]{@{}c@{}}tan./norm.\vspace{-4pt}\\ (mean)\end{tabular}}                                                                          & \multicolumn{2}{c|}{\begin{tabular}[c]{@{}c@{}}tan./norm.\vspace{-4pt}\\ (max)\end{tabular}}                                                                           & \multicolumn{1}{c|}{\multirow{2}{*}{\begin{tabular}[c]{@{}c@{}}loc.\vspace{-4pt}\\ error\vspace{-4pt}\\ (mm)\end{tabular}}} & \multicolumn{2}{c|}{\begin{tabular}[c]{@{}c@{}}tan./norm.\vspace{-4pt}\\ (mean)\end{tabular}}                                                                          & \multicolumn{2}{c||}{\begin{tabular}[c]{@{}c@{}}tan./norm.\vspace{-4pt}\\ (max)\end{tabular}}                                                      & \multicolumn{1}{c|}{\multirow{2}{*}{\begin{tabular}[c]{@{}c@{}}loc.\vspace{-4pt}\\ error\vspace{-4pt}\\ (mm)\end{tabular}}} & \multicolumn{2}{c|}{\begin{tabular}[c]{@{}c@{}}slipped\vspace{-4pt}\\ distance (mm)\end{tabular}}                                                                      & \multicolumn{1}{c|}{\multirow{2}{*}{\begin{tabular}[c]{@{}c@{}}loc.\vspace{-4pt}\\ error\vspace{-4pt}\\ (mm)\end{tabular}}} & \multicolumn{2}{c|}{\begin{tabular}[c]{@{}c@{}}slipped\vspace{-4pt}\\ distance (mm)\end{tabular}}                                                                      \\ \cline{3-6} \cline{8-11} \cline{13-14} \cline{16-17} 
                  & \multicolumn{1}{c|}{}                                                                             & \multicolumn{1}{c|}{\begin{tabular}[c]{@{}c@{}}small\vspace{-4pt}\\ motion\end{tabular}} & \multicolumn{1}{c|}{\begin{tabular}[c]{@{}c@{}}large\vspace{-4pt}\\ motion\end{tabular}} & \multicolumn{1}{c|}{\begin{tabular}[c]{@{}c@{}}small\vspace{-4pt}\\ motion\end{tabular}} & \multicolumn{1}{c|}{\begin{tabular}[c]{@{}c@{}}large\vspace{-4pt}\\ motion\end{tabular}} & \multicolumn{1}{c|}{}                                                                             & \multicolumn{1}{c|}{\begin{tabular}[c]{@{}c@{}}small\vspace{-4pt}\\ motion\end{tabular}} & \multicolumn{1}{c|}{\begin{tabular}[c]{@{}c@{}}large\vspace{-4pt}\\ motion\end{tabular}} & \multicolumn{1}{c|}{\begin{tabular}[c]{@{}c@{}}small\vspace{-4pt}\\ motion\end{tabular}} & \begin{tabular}[c]{@{}c@{}}large\vspace{-4pt}\\ motion\end{tabular} & \multicolumn{1}{c|}{}                                                                             & \multicolumn{1}{c|}{\begin{tabular}[c]{@{}c@{}}small\vspace{-4pt}\\ motion\end{tabular}} & \multicolumn{1}{c|}{\begin{tabular}[c]{@{}c@{}}large\vspace{-4pt}\\ motion\end{tabular}} & \multicolumn{1}{c|}{}                                                                             & \multicolumn{1}{c|}{\begin{tabular}[c]{@{}c@{}}small\vspace{-4pt}\\ motion\end{tabular}} & \multicolumn{1}{c|}{\begin{tabular}[c]{@{}c@{}}large\vspace{-4pt}\\ motion\end{tabular}} \\ \Xhline{2\arrayrulewidth}
proposed          & \multicolumn{1}{c|}{\textbf{0.89}}                                                                         & \multicolumn{1}{c|}{0.105}                                                  & \multicolumn{1}{c|}{0.096}                                                  & \multicolumn{1}{c|}{0.201}                                                  & \multicolumn{1}{c|}{\textbf{0.193}}                                                  & \multicolumn{1}{c|}{\textbf{0.81}}                                                                         & \multicolumn{1}{c|}{0.060}                                                  & \multicolumn{1}{c|}{0.091}                                                  & \multicolumn{1}{c|}{0.134}                                                  & \textbf{0.188}                                                  & \multicolumn{1}{c|}{\textbf{0.90}}                                                                         & \multicolumn{1}{c|}{0.19}                                                   & \multicolumn{1}{c|}{\textbf{0.35}}                                                   & \multicolumn{1}{c|}{\textbf{0.78}}                                                                         & \multicolumn{1}{c|}{0.06}                                                   & \multicolumn{1}{c|}{\textbf{0.16}}                                                   \\ \hline
constant tactile  & \multicolumn{1}{c|}{\textbf{0.94}}                                                                         & \multicolumn{1}{c|}{0.172}                                                  & \multicolumn{1}{c|}{0.176}                                                  & \multicolumn{1}{c|}{0.252}                                                  & \multicolumn{1}{c|}{\textbf{0.278}}                                                  & \multicolumn{1}{c|}{\textbf{1.30}}                                                                         & \multicolumn{1}{c|}{0.056}                                                  & \multicolumn{1}{c|}{0.132}                                                  & \multicolumn{1}{c|}{0.102}                                                  & \textbf{0.282}                                                  & \multicolumn{1}{c|}{\textbf{1.93}}                                                                         & \multicolumn{1}{c|}{0.84}                                                   & \multicolumn{1}{c|}{\textbf{12.90}}                                                  & \multicolumn{1}{c|}{\textbf{1.43}}                                                                         & \multicolumn{1}{c|}{0.06}                                                   & \multicolumn{1}{c|}{\textbf{4.27}}                                                   \\ \hline
w/o tactile E     & \multicolumn{1}{c|}{\textbf{1.42}}                                                                         & \multicolumn{1}{c|}{0.180}                                                  & \multicolumn{1}{c|}{0.216}                                                  & \multicolumn{1}{c|}{0.249}                                                  & \multicolumn{1}{c|}{\textbf{0.330}}                                                  & \multicolumn{1}{c|}{\textbf{1.38}}                                                                         & \multicolumn{1}{c|}{0.047}                                                  & \multicolumn{1}{c|}{0.156}                                                  & \multicolumn{1}{c|}{0.103}                                                  & \textbf{0.303}                                                  & \multicolumn{6}{c}{}                                                                                                                                                                                                                                                                                                                                                                                                                                                                                                          \\ \cline{1-11}
\end{tabular}
\vspace{-5mm}
\end{table*}

\begin{figure}[t]
	\centering
	\vspace{-1mm}
	\includegraphics[width=\linewidth]{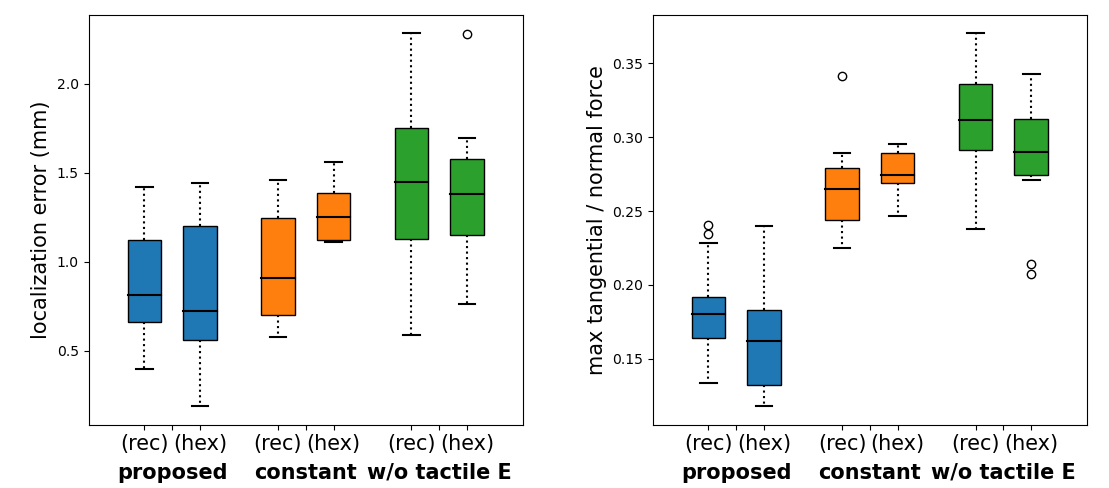}
	\vspace{-6mm}
	\caption{(left) Localization error, (right) maximum tangential to normal force ratio during the point contact motion on the higher friction surface}
	\label{fig:bar}
	\vspace{-7mm}
\end{figure}

\begin{figure}[t]
	\centering
	\includegraphics[width=0.8\linewidth]{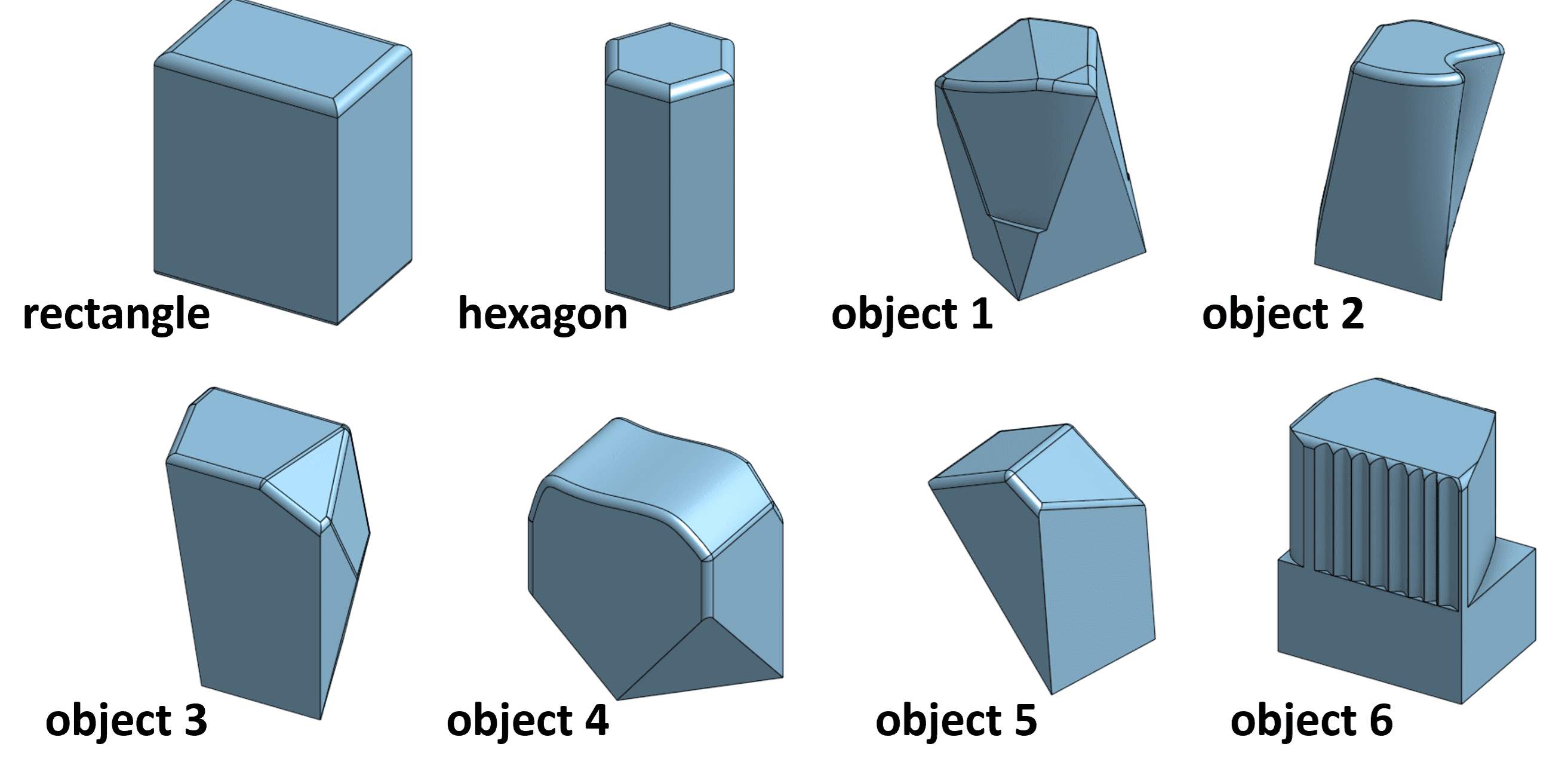}
	\vspace{-2mm}
	\caption{CAD model of the test objects}
	\label{fig:cad}
	\vspace{-3mm}
\end{figure}

Fig.\ref{fig:setup} shows the hardware setup for experiments. We use 6-DoF ABB's IRB120 robot arm and a WSG-50 gripper mounted with GelSlim 3.0 \cite{Taylor} on each side of the finger. ATI's Net Force-Torque sensor is mounted beneath the environment to collect the extrinsic contact force. We use eight 3D printed test objects, including rectangular and hexagonal cylinders and six others with irregular shapes (Fig.\ref{fig:cad}), to evaluate the generalization to various grasps and geometries. Pytorch \cite{Pytorch} is used for the tactile module computation, and iSAM2 \cite{Kaess2012} with GTSAM library \cite{Dellaert2012} is used to solve the least-squares problem incrementally.

\subsection{Point Contact Formation} \label{sec:line}

We test our method in a point contact formation to evaluate localization accuracy and capability to minimize the tangential contact force and slippage. We first grasp and orient the test objects with a random pose. Then we start a desired rotation command that draws a \textbf{small} conical spiral-like trajectory with a maximum of 5 degrees angle deviation from the initial orientation. Then we continue with a \textbf{large} cone-like desired rotation command with a maximum of 15 degrees from the initial orientation. We change the environment surface material between a higher friction paper, where we can evaluate the tangential force, and a lower friction acrylic, where we can evaluate the slippage. The result with 20 trials per each object is shown in Table.\ref{table:point} and Fig.\ref{fig:bar}. We show localization error and mean/max tangential to normal force ratio during the motion for the higher friction surface. For the lower friction surface, we show localization error with total slipped distance from the initial contact location. We report the result from two ablation models: the 'constant tactile' and the 'w/o tactile E.' The 'constant tactile' is the model in \cite{Kim2021}. It uses a proportional controller to keep the tactile displacement constant throughout the motion. In other words, it tries to maintain a constant intrinsic wrench. The 'w/o tactile E' is similar to our proposed method, but we make the weight of the tactile energy factor close to zero, so it will not prioritize minimizing the tactile energy.

\textbf{On the higher friction surface}, the proposed method showed approximately 25\% and 40\% reduction in localization error than the 'constant tactile' and 'w/o tactile E,' respectively. 
The difference is more significant in tangential to normal force ratio that the error bar of the proposed method does not overlap with those of ablation methods (Fig.\ref{fig:bar}) Hence, the proposed method is better at reducing the tangential forces required for pivoting, and less likely to slip.

\textbf{On the lower friction surface}, the proposed method is compared with the 'constant tactile' to evaluate the slip minimization capability. The 'constant tactile' showed small slippage with small motion but showed about two orders of magnitude higher slippage with large motion. This is because maintaining the constant intrinsic wrench will not be sufficient to keep the extrinsic contact force inside the friction cone if the motion is large. As a result of the large slip, the accuracy of localization also decreases. The 'constant tactile' shows about two times larger localization error than the proposed method.

\myparagraph{Contact Force Estimation}: From the estimated intrinsic wrench ($w$), we estimate the extrinsic contact force (e.g., dashed line in Fig.\ref{fig:plots}b) and evaluate the accuracy of the force estimation by comparing it with the F/T sensor measurement on the environment. However, the estimated force does not have a physical unit specified, so we only compare the mean difference in angle between the estimated force and the sensor measurement over all the data collected. To study the effect of incorporating the nonlinear term ($\Delta$) in Eq.\ref{eq:wrench}, we evaluate the accuracy without adding the nonlinear term. To study the effect of estimating the grasp parameters ($K$) rather than using initial prior, we use the initial prior values instead of estimated values and also remove the nonlinear term in Eq.\ref{eq:wrench}, then evaluate the accuracy. The result is shown in Table.\ref{table:force}. The performance gets better as we use the estimated grasp parameters instead of the initial prior values and also as the nonlinear term is considered. This implies that considering the nonlinear term and estimating the grasp parameters effectively improve the force estimation.

\subsection{Multiple Contact Formations}

\setlength{\tabcolsep}{6pt} 
\renewcommand{\arraystretch}{1.3}
\begin{table}[t]
\caption{Contact Force Estimation Accuracy}\label{table:force}
\vspace{-2mm}
\centering
\begin{tabular}{|c||cc|}
\hline
\multirow{2}{*}{}                                                                                & \multicolumn{2}{c|}{Mean Misalignment (deg)} \\ \cline{2-3} 
                                                                                                 & \multicolumn{1}{c|}{Rectangle}   & Hexagon   \\ \Xhline{2\arrayrulewidth}
proposed                                                                                         & \multicolumn{1}{c|}{6.2}         & 5.0       \\ \hline
w/o nonlinear term                                                                               & \multicolumn{1}{c|}{9.3}         & 13.0      \\ \hline
\begin{tabular}[c]{@{}c@{}}w/o nonlinear term \&\vspace{-4pt}\\ grasp parameters estimation\end{tabular} & \multicolumn{1}{c|}{13.9}        & 14.5      \\ \hline
\end{tabular}
\vspace{-6mm}
\end{table}

\begin{table}[t]
\caption{Evaluation of Simultaneous Tactile Estimator-Controller on Multiple Contact Formation}\label{table:multi}
\vspace{-2mm}
\centering
\begin{tabular}{|c||cc||c||ccc|}
\hline
\multirow{3}{*}{} & \multicolumn{2}{c||}{Point Contact}                                                                                                                                      & \multirow{3}{*}{\begin{tabular}[c]{@{}c@{}}Successful\\ Transition\\ Rate\end{tabular}} & \multicolumn{3}{c|}{Line Contact}                                                                                                                                                                                                              \\ \cline{2-3} \cline{5-7} 
                  & \multicolumn{1}{c|}{\multirow{2}{*}{\begin{tabular}[c]{@{}c@{}}loc.\vspace{-4pt}\\ err.\vspace{-4pt}\\ (mm)\end{tabular}}} & \multirow{2}{*}{\begin{tabular}[c]{@{}c@{}}slip\vspace{-4pt}\\ (mm)\end{tabular}} &                                                                                         & \multicolumn{1}{c|}{\multirow{2}{*}{\begin{tabular}[c]{@{}c@{}}loc.\vspace{-4pt}\\ err.\vspace{-4pt}\\ (deg)\end{tabular}}} & \multicolumn{2}{c|}{slip}                                                                                                                  \\ \cline{6-7} 
                  & \multicolumn{1}{c|}{}                                                                            &                                                                      &                                                                                         & \multicolumn{1}{c|}{}                                                                             & \multicolumn{1}{c|}{\begin{tabular}[c]{@{}c@{}}trans.\vspace{-4pt}\\ (mm)\end{tabular}} & \begin{tabular}[c]{@{}c@{}}rot.\vspace{-4pt}\\ (deg)\end{tabular} \\ \Xhline{2\arrayrulewidth}
rectangle         & \multicolumn{1}{c|}{0.80}                                                                        & 0.77                                                                 & 10 / 10                                                                                   & \multicolumn{1}{c|}{0.81}                                                                         & \multicolumn{1}{c|}{1.31}                                                       & 1.85                                                     \\ \hline
hexagon           & \multicolumn{1}{c|}{0.76}                                                                        & 0.48                                                                 & 8 / 10                                                                                    & \multicolumn{1}{c|}{4.54}                                                                         & \multicolumn{1}{c|}{0.84}                                                       & 1.16                                                     \\ \hline
object 1          & \multicolumn{1}{c|}{0.99}                                                                        & 0.63                                                                 & 8 / 10                                                                                    & \multicolumn{1}{c|}{2.66}                                                                         & \multicolumn{1}{c|}{1.73}                                                       & 1.75                                                     \\ \hline
object 2          & \multicolumn{1}{c|}{1.19}                                                                        & 0.87                                                                 & 6 / 10                                                                                    & \multicolumn{1}{c|}{4.24}                                                                         & \multicolumn{1}{c|}{3.41}                                                       & 1.98                                                     \\ \hline
object 3          & \multicolumn{1}{c|}{0.96}                                                                        & 0.94                                                                 & 10 / 10                                                                                   & \multicolumn{1}{c|}{1.58}                                                                         & \multicolumn{1}{c|}{1.18}                                                       & 1.48                                                     \\ \hline
object 4          & \multicolumn{1}{c|}{0.93}                                                                        & 1.08                                                                 & 9 / 10                                                                                    & \multicolumn{1}{c|}{0.82}                                                                         & \multicolumn{1}{c|}{1.12}                                                       & 1.39                                                     \\ \hline
object 5          & \multicolumn{1}{c|}{0.86}                                                                        & 0.47                                                                 & 8 / 10                                                                                    & \multicolumn{1}{c|}{2.03}                                                                         & \multicolumn{1}{c|}{1.25}                                                       & 1.70                                                     \\ \hline
object 6          & \multicolumn{1}{c|}{0.93}                                                                        & 0.75                                                                 & 10 / 10                                                                                   & \multicolumn{1}{c|}{1.84}                                                                         & \multicolumn{1}{c|}{0.83}                                                       & 1.00                                                     \\ \hline
\end{tabular}
\vspace{-6mm}
\end{table}

We test the proposed method on multiple contact formations, which requires the ability to detect the transition between different contact formations. Working in multiple contact formations is important in many manipulation tasks. 
Fig.\ref{fig:viz} and Fig.\ref{fig:plots} show an example of placing the rectangular object on a flat surface. It first makes initial point contact with the surface, executes motion to localize the point contact, detects the transition to the line contact, localizes the line contact, then finally detects the transition to the patch contact before releasing the grasp.

We run a similar experiments on the lower friction surface with all eight 3D printed objects. We focus on point and line contact because patch contact transition is trivial as it does not require or allow any motion after the transition. We first execute the same small desired rotation command as in Sec.\ref{sec:line}, then command the desired rotation to tilt roughly towards the direction where the object edge is. We assume we have a weak prior knowledge of the object edge's direction by adding a maximum error of 20 degrees to the true direction. After detecting the point-to-line contact transition, we modify the desired command rotation to draw a sinusoidal-like shape, where the sinusoid amplitude is perpendicular to the direction of tilt before the transition (dashed line in Fig.\ref{fig:plots}d). We estimate the line localization accuracy when the amount of rotation from the pose at the transition reaches 5 degrees.

Table.\ref{table:multi} shows the localization error, amount of slippage, and successful transition rate of the point-to-line contact transition. The failed transition includes the case where the transition is detected too early or late, so it loses the line contact, which leads to poor line estimation accuracy or causes the least-squares solver to throw an indeterminant error. The line contact evaluation metrics are calculated with only the successful cases.

Most test objects showed reasonably good localization errors, a small amount of slippage, and a high successful transition rate, while hexagon and object 2 showed relatively higher line localization errors. Object 2 also showed a lower successful transition rate and higher slippage amount than others. One main reason the hexagonal object showed higher line localization error is that it has a shorter edge than other objects. The hexagon edge is 17.5 mm, which is significantly shorter than other object edges; for example, the shortest edge of the rectangle is 35 mm. A short edge will make it harder for our method to detect the point-to-line contact transition, and it is more likely that the object loses the line contact with the surface during  motion. Object 2's worse performance is likely due to the different local grasp geometry. It has a high curvature and sharp cut at the grasped part, which leads to a smaller contact patch with the fingers. This might makes the grasp parameters ($K$) deviate much from that of other objects and lead to worse performance.

%

\section{Conclusion and Future Work}

We demonstrate simultaneous estimation and control for contact state of unknown objects using tactile sensing. We show that we can localize the extrinsic contact by fusing the gripper pose and tactile measurements in a factor graph. Regressing the grasp mechanics to a decoupled stiffness model of the grasp enables the estimation of the grasp wrench and the contact force. Also, we impose control objectives as factors in the same graph to execute the pivoting motion with minimal tangential force and slip. Lastly, we implement the method on multiple contact formations and are able to successfully detect the transition between them.

While the proposed work focuses only on minimizing the tangential forces on the environment surface, one possible follow-up is reasoning about normal force. A trade-off is that a larger normal force will enhance observability but possibly stress the environment and object. Extending the proposed work to not only the sticking contact but other contact modes is also an important direction of research. Lastly, we plan to develop an automated desired command rotation policy instead of manually commanding it.



\bibliographystyle{IEEEtran}
\bibliography{ICRASK}

\end{document}